\documentclass[a4paper]{amia}
\usepackage{lipsum} %Remove if not needed

\begin{document}

\title{Explorers at \#SMM4H 2023: Enhancing BERT for Health Applications through Knowledge and Model Fusion}

\author{Xutong Yue$^{1,\dag}$, Xilai Wang$^{2,\dag}$, Yuxin He$^1$, Zhenkun Zhou$^1$}

\institutes{
    $^1$ School of Statistics, Capital University of Economics and Business, Beijing, China\\
    $^2$ School of EECS, University of Ottawa, Ottawa, Canada\\
    $^{\dag}$ These authors contributed equally.\\
    % $^*$ Corresponding author    
}

\maketitle
\begin{center}
\texttt{yuexutong@gmail.com, xwang736@uottawa.ca, \{heyuxin, zhenkun\}@cueb.edu.cn} \\
\end{center}
\section*{Abstract}
% \vspace{-\baselineskip}

An increasing number of individuals are willing to post states and opinions in social media, which has become a valuable data resource for studying human health. Furthermore, social media has been a crucial research point for healthcare now. This paper outlines the methods in our participation in the \#SMM4H 2023 Shared Tasks, including data preprocessing, continual pre-training and fine-tuned optimization strategies. Especially for the Named Entity Recognition (NER) task, we utilize the model architecture named W2NER\cite{li2022unified} that effectively enhances the model generalization ability. Our method achieved first place in the Task 3. \footnote{This paper has been peer-reviewed and accepted for presentation at the \#SMM4H 2023 Workshop.}

\section{Introduction}

The Social Media Mining for Health Applications (SMM4H) Shared Tasks aim to address natural language processing (NLP) challenges of using social media data for health informatics, including informal, colloquial expressions, misspellings, noise, data sparsity, ambiguity, and multilingual posts\textsuperscript{~\cite{smm4h}}
. We participated in three tasks in SMM4H 2023 (i.e. Task 1, Task 3, Task 4). Task 1 and Task 4 focus on COVID-19 diagnosis in self-reported English tweets and self-reported social anxiety disorder diagnosis posted in Reddit. Task 3 concentrates on detecting and extracting COVID-19 symptoms in Latin American Spanish tweets description.\par
The dataset of Task 1 contain texts from Twitter that self-report the COVID-19 diagnosis (labeled as `1') or not (labeled as `0'). The size of the training set, validation set, and test set are 7600, 400, 10000. Task 4 contains 8117 posts from users aged 12 to 25. Positive cases (labeled as `1') represent self-reported or probable social anxiety disorder diagnoses, while negative cases (labeled as `0') include users without a diagnosis or with uncertain diagnostic status. The sizes of the training set, validation set, and those of test set are 6090, 680, 1347. Task 3 focuses on the detection and extraction of COVID-19 symptoms in tweets written specifically in Latin American Spanish, includes both personal self-reports and third-party mentions of symptoms. There are 6021 of the training data, 1979 for validation, and 2150 for testing.

In the paper, we present methodologies for addressing three tasks. For the classification tasks, we adopt three pre-trained language models (PLMs) as our backbones: RoBERTa-base\textsuperscript{~\cite{Conneau2019UnsupervisedCR}}, vinai-bertweet\textsuperscript{~\cite{bertweet}}, and CPM-RoBERTa. For NER task, we select bert-base-spanish-wwm-cased-xnli\textsuperscript{~\cite{xnli}}, BETO\textsuperscript{~\cite{canete2023spanish}}, and BETO\_NER\textsuperscript{~\cite{BETO_NER}}. We also propose tricks to mitigate the challenges existing in dataset, including category imbalance, domain disparities, and text noise. Our approach achieves state-of-the-art results in Task 3 and surpasses the median performance in Task 1.

\section{Methods}
Methods used for task 1 and 4 consist of three main components: continual pre-processing, continual pre-training, and model fine-tuning. In Task 3, based on methods for Task 1 and 4, we use additiaonally the W2NER model architecture.\par
%To achieve optimal performance,  In Task 3, we adopted three PLMs: , bert-base-spanish-wwm-cased-finetuned-ner, and spanish\_wwm\_cased as backbones. During the training phase, we also implemented various strategies to improve the overall score. Details are provided below.\par
% \vspace{-\baselineskip}
\paragraph{Pre-processing}
% \subsection*{Pre-processing}
% \vspace{-\baselineskip}
% \vspace{1em}
In the provided dataset, it contains a large number of emojis and usernames, which lacks valid textual semantic information. To enhance the model's focus on textual semantics, we preprocess the data by replacing emojis with corresponding emotion description, and usernames with `@user'. Hashtags contains some information sometimes, we preprocess `\#hashtag' by separating `\#' and the text follows.
% \vspace{-\baselineskip}
\paragraph{Continual Pre-training}
% \subsection*{Continual Pre-training }
% \vspace{1em}
% \vspace{-\baselineskip}
The purpose of continual pre-training is to obtain pre-trained models that are more applicable to specific industries or domains, aiming to improve the effectiveness of the models in downstream tasks\textsuperscript{~\cite{gururangan2020don}}. Pretrained models are usually trained on large-scale generalized data, and thus may not have a deep enough understanding of some task-specific data. To solve the domain bias problem, we continue with domain-adaptive pretraining using unlabeled data from all the related tasks to optimize the model for task-specific data\textsuperscript{~\cite{fu2022casia}}. Specifically, for the classification task, we select all relevant COVID-19 self-reported English tweets data, and anxiety self-reported data to continue the task-adaptive pretraining of the three pretrained models with a learning rate of 3e-5 and 20 epochs. For the NER task, we use the training and validation data for continual pre-training, with the learning rate of 3e-5 and 30 epochs.
The effects of continual pre-training are quite unstable, sometimes leading to positive impacts and other times resulting in negative influences, which are largely correlated with the data used for pre-training. Fortunately, in the 3 tasks we conducted, continual pre-training yielded positive results.
% \vspace{-\baselineskip}
\paragraph{W2NER Model Architecture}

% \subsection*{W2NER Model Architecture}
% \vspace{-\baselineskip}
In Task 3, rather than employing a conventional model architecture, we adopt the state-of-the-art W2NER model architecture. The W2NER architecture captures neighboring relations between entity words using two distinct types of relations(the Next-Neighboring-Word, NNW, and the Tail-Head-Word, THW). Building upon this approach, the creator of the W2NER model introduced a neural framework that represents unified Named Entity Recognition (NER) as a 2D grid of word pairs. Leveraging the W2NER architecture, we achieved the top ranking in Task 3.
% \vspace{-\baselineskip}
\paragraph{Optimization Strategy}

% \subsection*{Optimization Strategy}
% \vspace{-\baselineskip}

In Task 1 and 4, we implemented all 4 optimization strategies below. For Task 3, only Multi-Model Fusion Voting is employed.\par
\noindent
\textbf{Focal Loss:} Classification tasks exhibit noticeable category imbalance. The Task 1 has a positive example proportion of 17.6\%, and the Task 4 has 37.8\%. To address the problem, we utilize focal loss to adjust sample weights during training, emphasizing hard samples\textsuperscript{~\cite{lin2017focal}}. Specifically, for binary classification, the traditional cross-entropy loss function is expressed as:
\begin{equation}
CE(p,y)=-(y \cdot log(p) + (1-y) \cdot log(1-p))
\end{equation}
Focal loss introduces a modifying factor $\gamma$. The modified loss function is:
\begin{equation}
FL(p,y)=-\alpha \cdot (1-p)^\gamma \cdot log(p)
\end{equation}
We also tried to randomly replicate samples in a few classes using oversampling to make the training data more balanced between positive and negative samples. \par
\noindent
\textbf{Single-model Fusion Voting:} In Task 1 and 4, we employ five-fold cross-validation, dividing the training set into five subsets. Four subsets are used for training, and one is for development. Using the same model structure and initial parameters, we train five models on the subsets. The models' results are combined through average voting and applied to the test data. \par
\noindent
\textbf{Parameter Average Generalization:} Overfitting occurs when a model learns excessive details and noises from the training data, resulting in decreased generalization performance on the real data. To enhance the model's robustness and reduce training data oscillations caused by noise and variations, we employ parameter average generalization. 
% This approach smooths the model's parameters to mitigate overfitting.Common methods include:
Specifically, exponential sliding average (EMA): $\theta_{t}$ is the model weight at time $t$.The original weights are still used during the training process and shadow weights are calculated after a certain number of steps, which are used in the testing phase to replace the weights in the neural network: 
% \begin{itemize}
% \item Stochastic sliding average (SWA): at a later stage of the training process, the model parameters are averaged multiple times, i.e., the model parameters of multiple cycles are averaged to obtain an average model parameter.
\begin{equation}
v_{n}=\alpha \cdot v_{n-1} + (1-\alpha)\theta_{n}
\end{equation}
\begin{equation}
v_{t}=\frac{v_{t}}{(1-\beta^t)}
\end{equation}
% \end{itemize}
\noindent
\textbf{Multi-Model Fusion Voting:} In order to further improve the model generalization ability and achieve the complementary advantages of different models under the unified task, we use multi-model fusion voting in Task 1, 3 and 4 to get the final prediction results of different model prediction results for the same task through Mean-Pooling. For Task 3, bert-base-spanish-wwm-cased-xnli, BETO, and BETO\_NER are used for Multi-Model Fusion Voting.\par
\noindent
\section{Results}
\paragraph{Task 1 and Task 4}
% \vspace{-\baselineskip}

For each model, we set batch size to 32, epoch to 30, learning rate to 3e-5 and weight decay to 0.01 for bert's internal parameters, set learning rate to 3e-4 and weight decay to 0 for the classification layer parameters. the optimizer uses AdamW\textsuperscript{\cite{kingma2014adam}}. For the unbalanced samples, we use focal loss with a weight parameter of 0.25, and the adjustment factor of 2. The experiments’ results on validation are shown in Table 1.
Continual pre-training is beneficial to allow the pre-trained model to learn some domain-specific information, in order to test the effectiveness of continual pre-training to improve the performance of the model, we present the results of two versions, our F1-scores are shown in Table 2.
For each task, we present model predictions that show the best performance in the validation set, our model exceeds the mean of Task 1 by 0.06. The results on Task 4 are below the median.

\vspace{1em}

\begin{table}[H]
\begin{center}
    % \centerline{\parbox{13.8cm}{Table 1:F1-scores of classification tasks. Original means using the original training dataset, Oversampling means using the oversampled set, and Focal\_Loss means using only focal\_loss without oversampling.}}
\caption{Table 1:F1-scores of classification tasks. Original means using the original training dataset, Oversampling means using the oversampled set, and Focal\_Loss means using only focal\_loss without oversampling.}
\begin{tabular}{|l|c|c|c|}
\hline
Task  & Original  & Oversampling  & Focal Loss\\
\hline
\#1   & 0.806    & 0.885   & 0.870\\
\hline 
\#4   & 0.648    & 0.687   & 0.709\\
\hline
\end{tabular}
\end{center}
\label{tab:table1}
\end{table}

\begin{table}[H]
\begin{center}
\caption{F1-scores of classification tasks, Original means that Continual Pre-training is not used.}
\begin{tabular}{|l|c|c|}
\hline
Task & Original & Continual Pre-training\\
\hline
\#1   &  0.865   &  0.872   \\
\hline 
\#4   &  0.677   &  0.695   \\
\hline
\end{tabular}
\end{center}
\label{tab:submission}
\end{table}

% \vspace{-\baselineskip}
\paragraph{Task 3}
% \vspace{-\baselineskip}
Regarding each model, we configured the batch size as 12, the number of epochs as 30, and the learning rate for BERT's internal parameters as 1e-5. Additionally, we designated a learning rate of 1e-3 and a weight decay of 5e-3 for the remaining parameters. 
We also employ continual pre-training and multi-model fusion voting in Task 3. For the continual pre-training, the number of epochs is 30, learning rate is 3e-5.
As a result, on the test set, our precision is 0.94, recall is 0.93, F1-score is 0.94. Our F1-score is the highest among all the submissions for Task 3.

\section{Conclusion}
The SMM4H 2023 Shared Tasks aim to extract and understand users' opinions about healthcare from social media. We propose a deep learning model to tackles category imbalance, domain bias and text noise. Due to time constraints, we were unable to address all issues, such as misspellings and informal or colloquial expressions in the Task 3. For instance, during symptom recognition, the model successfully identifies the term `gripe; but overlooks variations like `gripes' and `gripa'. Future research that successfully resolves this challenge has the potential to achieve the model's performance.

\section*{Acknowledgements}
Xutong Yue, Yuxin He and Zhenkun Zhou were supported by the R\&D Program of Beijing Municipal Education Commission, Grant No. KM202210038002.

% References as numbers

%\makeatletter
%\renewcommand{\@biblabel}[1]{\hfill #1.}
%\makeatother
\newpage
% unstr is used to keep citation order
\bibliographystyle{vancouver}

\begin{thebibliography}{10}

\bibitem{li2022unified}
Li J, Fei H, Liu J, Wu S, Zhang M, Teng C, et~al.
\newblock Unified named entity recognition as word-word relation
  classification.
\newblock In: Proceedings of the AAAI Conference on Artificial Intelligence.
  vol.~36; 2022. p. 10965-73.

\bibitem{smm4h}
{Klein AZ, Banda JM, Guo Y, Flores Amaro JI, Rodriguez-Esteban R, Sarker A,
  Schmidt AL, Xu D, Gonzalez-Hernandez G}.
\newblock {Overview of the eighth Social Media Mining for Health Applications
  (\#SMM4H) Shared Tasks}.
\newblock Proceedings of the Eighth Social Media Mining for Health Applications
  (\#SMM4H) Workshop and Shared Task. 2023.

\bibitem{Conneau2019UnsupervisedCR}
Conneau A, Khandelwal K, Goyal N, Chaudhary V, Wenzek G, Guzm{\'a}n F, et~al.
\newblock Unsupervised Cross-lingual Representation Learning at Scale.
\newblock In: Annual Meeting of the Association for Computational Linguistics;
  2019. 
  
%  Available from:
%  \url{https://api.semanticscholar.org/CorpusID:207880568}.

\bibitem{bertweet}
Nguyen DQ, Vu T, Nguyen AT.
\newblock {BERTweet: A pre-trained language model for English Tweets}.
\newblock In: Proceedings of the 2020 Conference on Empirical Methods in
  Natural Language Processing: System Demonstrations; 2020. p. 9-14.

\bibitem{xnli}
Recognai. bert-base-spanish-wwm-cased-xnli; 2021.
%\newblock Available from:
%  \url{https://huggingface.co/Recognai/bert-base-spanish-wwm-cased-xnli}.

\bibitem{canete2023spanish}
Ca{\~n}ete J, Chaperon G, Fuentes R, Ho JH, Kang H, P{\'e}rez J.
\newblock Spanish pre-trained bert model and evaluation data.
\newblock arXiv preprint arXiv:230802976. 2023.

\bibitem{BETO_NER}
Romero M. bert-spanish-cased-finetuned-ner; 2020.
%\newblock Available from:
%  \url{https://huggingface.co/mrm8488/bert-spanish-cased-finetuned-ner}.

\bibitem{gururangan2020don}
Gururangan S, Marasovi{\'c} A, Swayamdipta S, Lo K, Beltagy I, Downey D, et~al.
\newblock Don't stop pretraining: Adapt language models to domains and tasks.
\newblock arXiv preprint arXiv:200410964. 2020.

\bibitem{fu2022casia}
Fu J, Gan Z, Li Z, Li S, Sui D, Chen Y, et~al.
\newblock CASIA at SemEval-2022 Task 11: Chinese named entity recognition for
  complex and ambiguous entities.
\newblock In: Proceedings of the 16th international workshop on semantic
  evaluation (SemEval-2022); 2022. p. 1518-23.

\bibitem{lin2017focal}
Lin TY, Goyal P, Girshick R, He K, Doll{\'a}r P.
\newblock Focal loss for dense object detection.
\newblock In: Proceedings of the IEEE international conference on computer
  vision; 2017. p. 2980-8.

\bibitem{kingma2014adam}
Kingma DP, Ba J.
\newblock Adam: A method for stochastic optimization.
\newblock arXiv preprint arXiv:14126980. 2014.

\end{thebibliography}

\end{document}